\documentclass{article}

\usepackage{arxiv}

\usepackage[utf8]{inputenc} % allow utf-8 input
\usepackage[T1]{fontenc}    % use 8-bit T1 fonts
\usepackage{hyperref}       % hyperlinks
\usepackage{url}            % simple URL typesetting
\usepackage{booktabs}       % professional-quality tables
\usepackage{amsfonts}       % blackboard math symbols
\usepackage{nicefrac}       % compact symbols for 1/2, etc.
\usepackage{microtype}      % microtypography
\usepackage{lipsum}
\usepackage{graphicx}
\graphicspath{ {./images/} }

\title{Kinematics of motion tracking using computer vision}

\author{
 José L. Escalona \\
  Dept. of Mechanical and Manufacturing Engineering\\
  University of Seville\\
  \texttt{escalona@us.es} \\
}

\begin{document}
\maketitle
\begin{abstract}
This paper describes the kinematics of the motion tracking of a rigid body using video recording. The novelty of the paper is on the adaptation of the methods and nomenclature used in Computer Vision to those used in Multibody System Dynamics. That way, the equations presented here can be used, for example, for inverse-dynamics multibody simulations driven by the motion tracking of selected bodies. This paper also adapts the well-known Zhang calibration method to the presented nomenclature.
\end{abstract}

% keywords can be removed

\section{Introduction}
This document describes the kinematic relations used to find the motion of a body using video recording. The theory presented here is well known in Computer Vision technology \cite{szelinski2011computerVision}. The nomenclature is adapted here to match that used in Multibody System Dynamics.\\

\section{Kinematics of the video camera} \label{sec:kinematicsVideoCamera}
\begin{figure}[htbp!]
	\centering
	\includegraphics[width=0.6\linewidth]{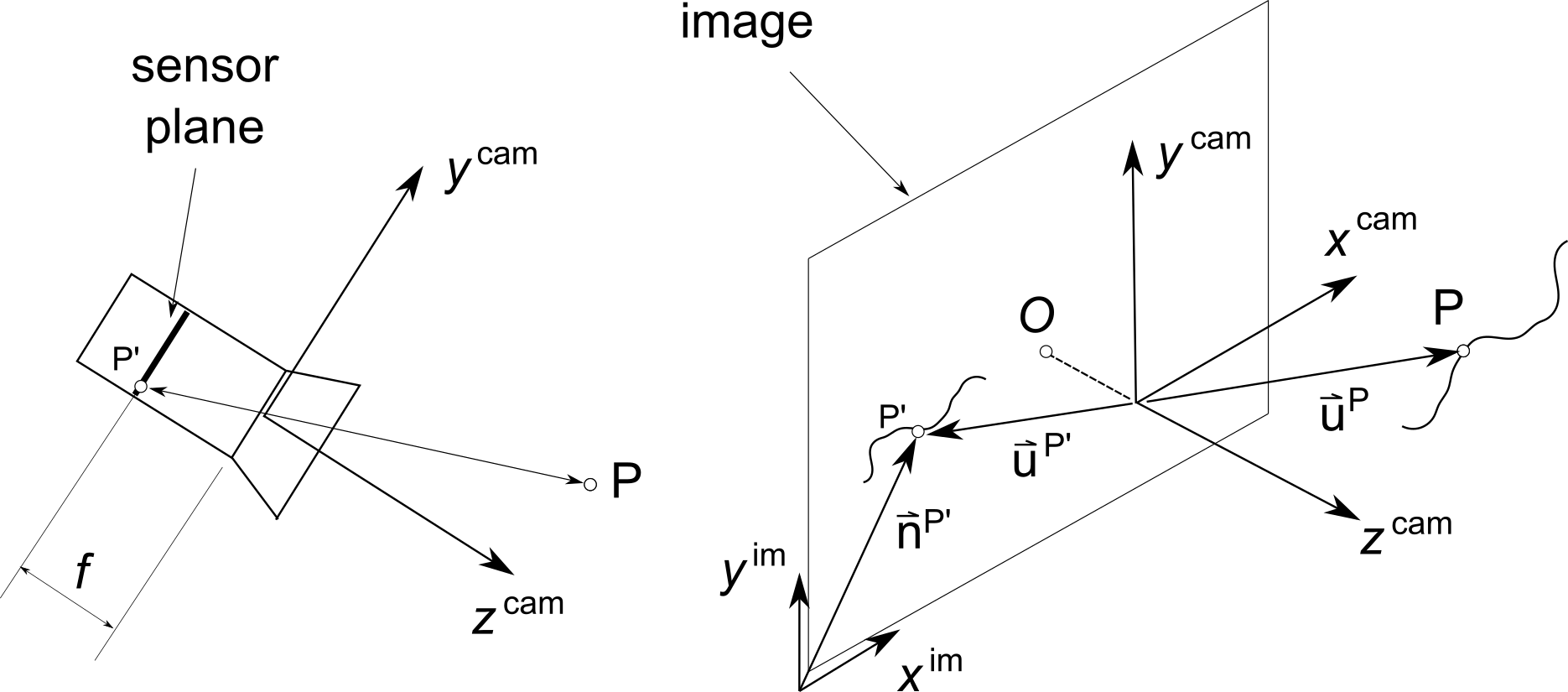}
	\caption{Position vectors in camera frame and image frame}
	\label{fig:cameraPositionVectors}
\end{figure}

Figure \ref{fig:cameraPositionVectors} on the left shows a schematic sketch of a video camera. The theory presented here is called the pinhole model of the camera. It is inspired in the photo cameras that were used when photography was invented. Light entered into the camera through a hole and the image was formed on the back face of a dark box were a photosensitive plate was located. Nowadays cameras do not work that way but they use lenses. However, the pinhole model is still used to relate the position of points in the world to the position of the point in the image. 

The frame $\left\langle {{x^{cam}},\,\,{y^{cam}},\,\,{z^{cam}}} \right\rangle $ shown in Fig. \ref{fig:cameraPositionVectors} is called the \textit{camera frame}. The origin is assumed to be located at the pinhole. Axis $z^{cam}$, that is assumed to be perpendicular to the sensor plane (where the image is formed), is called optical axis. The distance from the pinhole to the sensor plane f is called focal length. The origin of the sensor frame $\left\langle {{x^{cam}},\,\,{y^{cam}}} \right\rangle $ is located at the right (when watched from behind, where the cameraman would be located) –down corner of the image. However, because images are formed upside-down in the sensor, the sensor frame appears in the left-up corner when watching the image on a screen.

Assume a point $P$ on the real world that appears as point $P’$ on the image (see Fig. \ref{fig:cameraPositionVectors} on the right). The position vector of both points in the camera frame are given by the following components:

\begin{equation} \label{uPuPprima}
{{\bf{\bar u}}^P} = \left[ {\begin{array}{*{20}{c}}
	{\bar u_x^P}\\
	{\bar u_y^P}\\
	{\bar u_z^P}
	\end{array}} \right],\,\,\,\,\,\,\,\,{{\bf{\bar u}}^{P'}} = \left[ {\begin{array}{*{20}{c}}
	{\bar u_x^{P'}}\\
	{\bar u_y^{P'}}\\
	{ - f}
	\end{array}} \right]
\end{equation}

where the bar over the symbol means that the components are given in the camera frame (not in the world frame that will be introduced later). As it can be observed in Fig. \ref{fig:cameraPositionVectors} on the left, clearly, the value of $\bar u_z^{P'}$ is always –f. According to the pinhole model, these two positon vectors, ${{\bf{\bar u}}^P}$   and ${{\bf{\bar u}}^{P'}}$, are related. They are collinear and their components fulfill the following equations:

\begin{equation} \label{Thales}
\frac{{\bar u_x^P}}{{\bar u_x^{P'}}} = \frac{{\bar u_y^P}}{{\bar u_y^{P'}}} = \frac{{\bar u_z^P}}{{ - f}}
\end{equation}

The reason behind these equations is Thales theorem applied to the similar triangles that the position vectors form with their components. Equation \ref{Thales} can be written in vector form as:

\begin{equation} \label{ThalesMatrix}
{{\bf{\bar u}}^{P'}} = \frac{{ - f}}{{\bar u_z^P}}{{\bf{\bar u}}^P}
\end{equation}

\begin{figure}[ht]
\begin{center}
	\includegraphics[width=0.6\textwidth]{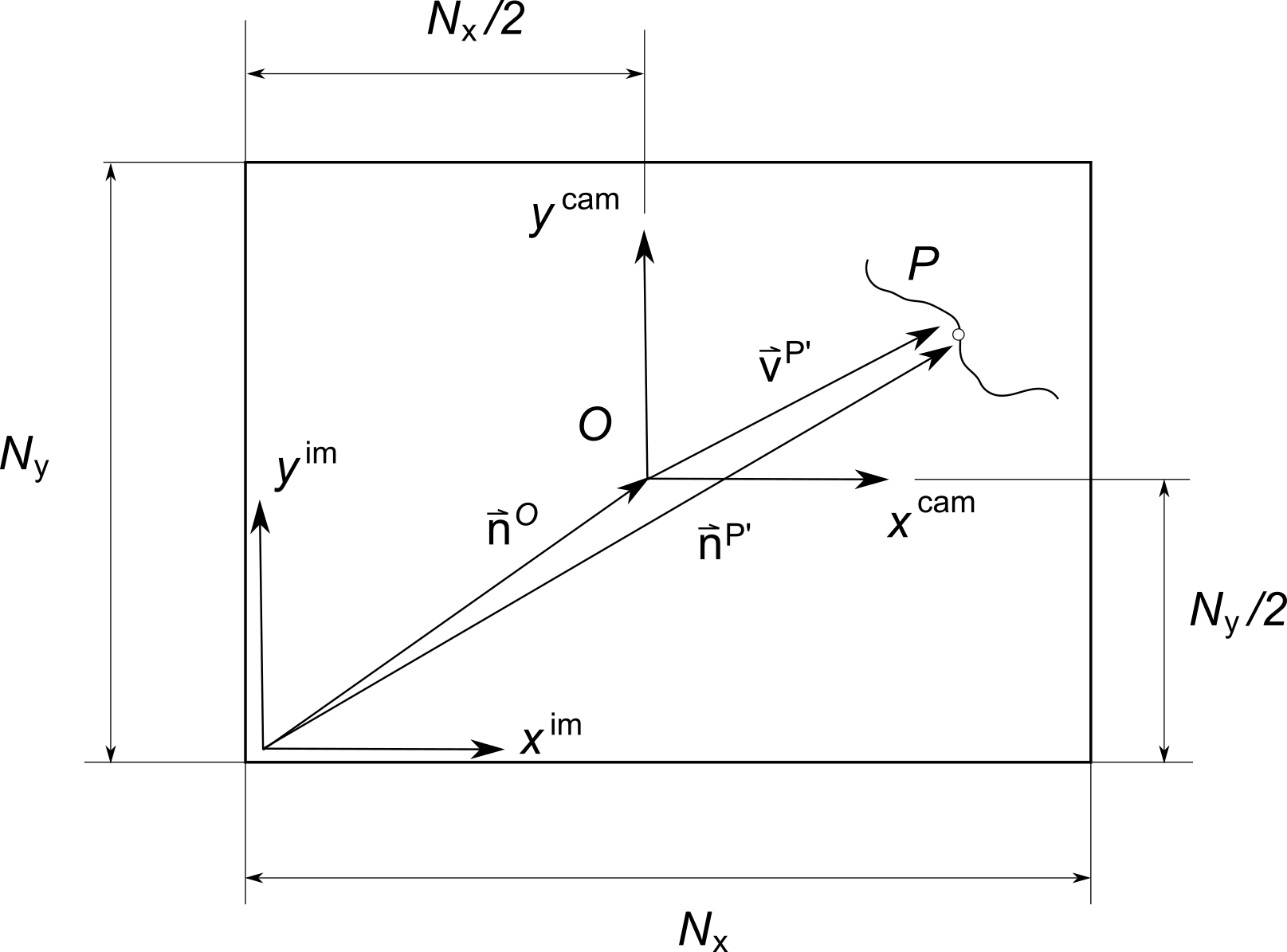}
	\caption{Frames in sensor plane}
    \label{fig:sensorPlane}
    \end{center}
\end{figure}

Figure \ref{fig:sensorPlane} represents the image or sensor plane. The components of position vectors in the image ${{\bf{n}}^{P'}}$ (2D vector) are not measured in units of length (meters or millimeters) but in pixels (non-dimensional). Vector ${{\bf{\bar v}}^{P'}}$ is the 2D projection of ${{\bf{\bar u}}^{P'}}$ in the image plane (units in meters). Clearly, ${{\bf{n}}^{P'}}$ and ${{\bf{\bar v}}^{P'}}$ can be related if two assumptions are made:

\begin{enumerate}
	\item That the optical axis $z^{cam}$ is perpendicular to the sensor plane and $x^{cam}$ and $y^{cam}$ are parallel to $x^{im}$ and $y^{im}$, respectively, and
	\item The optical axis $z^{cam}$ intersects the sensor plane exactly in the center, such that the position of that intersection point $O$ in the sensor plane is given by:
	\begin{equation} \label{nO}
	{{\bf{n}}^O} = \left[ {\begin{array}{*{20}{c}}
		{{{{N_x}} \mathord{\left/
					{\vphantom {{{N_x}} 2}} \right.
					\kern-\nulldelimiterspace} 2}}\\
		{{{{N_y}} \mathord{\left/
					{\vphantom {{{N_y}} 2}} \right.
					\kern-\nulldelimiterspace} 2}}
		\end{array}} \right]
	\end{equation}
	being $N_x$ and $N_y$ the total number of pixels in the x and y directions.
	
\end{enumerate}

Under these conditions, vector ${{\bf{\bar v}}^{P'}}$ contains the \textit{x} and \textit{y} components of ${{\bf{\bar u}}^{P'}}$, as follows:

\begin{equation} \label{vPprima}
{{\bf{\bar v}}^{P'}} = \left[ {\begin{array}{*{20}{c}}
	{\bar u_x^{P'}}\\
	{\bar u_y^{P'}}
	\end{array}} \right]
\end{equation}

It is easy to find that:

\begin{equation} \label{nPprima}
{{\bf{n}}^{P'}} = {{\bf{n}}^O} + \frac{1}{s}{{\bf{\bar v}}^{P'}} = {{\bf{n}}^O} - \frac{f}{{s\bar u_z^P}}{{\bf{\bar u}}^P}
\end{equation}

where \textit{s} is a constant that gives the size of the width or height of the pixels in meters (in some cameras the size of the width and height of the pixels can be different, however, without loss of generality, we will initially assume them to be equal) and the relations given in Eqs. \ref{ThalesMatrix} and \ref{vPprima} have been used.

\section{Homogeneous coordinates of a vector}  \label{sec:Homogeneous}

At this point, it is convenient to introduce the concept of the \textit{homogeneous coordinates of a vector} that are commonly used in the theory of computer vision. A vector, that can be 2D or 3D, is represented by 3 or 4 components, respectively, when represented by homogeneous coordinates, as follows:

\begin{equation} \label{hom2D}
{\bf{v}} = \left[ {\begin{array}{*{20}{c}}
{{v_x}}\\
{{v_y}}
\end{array}} \right]\,\,\,\, \Rightarrow \,\,\,\,\,{\grave{\bf{v}}} = \left[ {\begin{array}{*{20}{c}}
{a{v_x}}\\
{a{v_y}}\\
a
\end{array}} \right],\,\,\,\,\,{\grave{\bf{v}}} = \hom ({\bf{v}}),\,\,\,{\bf{v}} = {\hom ^{ - 1}}(\grave{\bf{v}})
\end{equation}

\begin{equation} \label{hom3D}
{\bf{w}} = \left[ {\begin{array}{*{20}{c}}
{{w_x}}\\
{{w_y}}\\
{{w_z}}
\end{array}} \right]\,\,\, \Rightarrow \,\,\,\,\,\grave{\bf{w}} = \left[ {\begin{array}{*{20}{c}}
{b{w_x}}\\
{b{w_y}}\\
{b{w_z}}\\
b
\end{array}} \right],\,\,\,\,\,\grave{\bf{w}} = \hom ({\bf{w}}),\,\,\,{\bf{w}} = {\hom ^{ - 1}}(\grave{\bf{w}})
\end{equation}

where $\grave{\bf{v}}$ and $\grave{\bf{w}}$ are the homogeneous versions of ${\bf{v}}$ and ${\bf{w}}$ and $a$ and $b$ are non-zero arbitrary real numbers. In order to get  or  from their homogeneous versions, one just needs to divide the two first or three first components, respectively, by the last one. Clearly, the real numbers $a$ and $b$ are irrelevant, such that the following homogeneous vectors are equivalent:

\begin{equation} \label{v1v2}
\grave{\bf{v}}_1 = \left[ {\begin{array}{*{20}{c}}
{a{v_x}}\\
{a{v_y}}\\
a
\end{array}} \right] \equiv \left[ {\begin{array}{*{20}{c}}
{c{v_x}}\\
{c{v_y}}\\
c
\end{array}} \right] = \grave{\bf{v}}_2
\end{equation}

this is, they are homogeneous representations of the same 2D vector $\bf{v}$. 

\section{Intrinsic parameters of the camera}  \label{sec:Intrinsics}
Using the homogeneous representation of vector ${{\bf{n}}^{P'}}$ , Eq. \ref{nPprima} can be written as: 

\begin{equation} \label{nPprima2}
{\grave{\bf{n}}^{P'}} = \left[ {\begin{array}{*{20}{c}}
{{{ - f} \mathord{\left/
 {\vphantom {{ - f} s}} \right.
 \kern-\nulldelimiterspace} s}}&0&{{{{N_x}} \mathord{\left/
 {\vphantom {{{N_x}} 2}} \right.
 \kern-\nulldelimiterspace} 2}}\\
0&{{{ - f} \mathord{\left/
 {\vphantom {{ - f} s}} \right.
 \kern-\nulldelimiterspace} s}}&{{{{N_y}} \mathord{\left/
 {\vphantom {{{N_y}} 2}} \right.
 \kern-\nulldelimiterspace} 2}}\\
0&0&1
\end{array}} \right]\left[ {\begin{array}{*{20}{c}}
{\bar u_x^P}\\
{\bar u_y^P}\\
{\bar u_z^P}
\end{array}} \right] = {{\bf{M}}^{int}}{{\bf{\bar u}}^P}
\end{equation}

where the $3\times3$ matrix ${{\bf{M}}^{int}}$ is called \textit{matrix of intrinsic parameters} of the camera and it relates the homogeneous representation of the position vector of a point on the image to its position vector in the camera frame. The intrinsic parameters of the camera are $f$, $s$, $N_x$ and $N_y$. 

Equation \ref{nPprima2} matches the position of the points in the image  ${{\bf{n}}^{P'}}$ to their position in the real world ${{\bf{\bar u}}^P}$ . This equation can be used just in one direction: to obtain the position in the camera once the position in the world is known. If the value of ${{\bf{\bar u}}^P}$ is known, Eq. \ref{nPprima2} is applied and the operation ${{\bf{n}}^{P'}} = {\hom ^{ - 1}}\left( {{{\grave{\bf{n}}}^{P'}}} \right)$ , that is just to divide the two first components by the third one, provides the position in the image. However, it is not possible to get the position of the point in the world ${{\bf{\bar u}}^P}$ once the position in the camera  ${{\bf{n}}^{P'}}$ is known. To that end, the value of the distance of the point to the camera $\bar u_z^P$ would be needed. This result is consistent with the idea that the 3D world cannot be reconstructed from a 2D projection of it.

The definition of the elements of ${{\bf{M}}^{int}}$ is given by the intrinsic parameters only in the case that the assumptions made previously (optical axis perpendicular to the sensor plane in the midpoint) are valid. In general, these assumptions can be relaxed. In such a case, the matrix of intrinsic parameters is defined as: 

\begin{equation} \label{Mint2}
{{\bf{M}}^{int}} = \left[ {\begin{array}{*{20}{c}}
\alpha &\gamma &{n_x^O}\\
0&\beta &{n_y^O}\\
0&0&1
\end{array}} \right]
\end{equation}

that has the same form as the one shown in Eq. \ref{nPprima2}, with the exception of the new non-zero term $\gamma$  that accounts for the skewness of the two image axis. Therefore, finding the intrinsic parameters of the camera consists on finding the values of $\alpha ,\,\beta ,\,\,n_x^O,\,n_y^O\,{\rm{and}}\,\gamma $. The first four of them are related to $f$, $s$, $N_x$ and $N_y$ as shown in Eq. \ref{nPprima2}.

\section{Extrinsic parameters of the camera}  \label{sec:Extrinsics}

\begin{figure}[htbp!]
	\centering
	\includegraphics[width=0.4\linewidth]{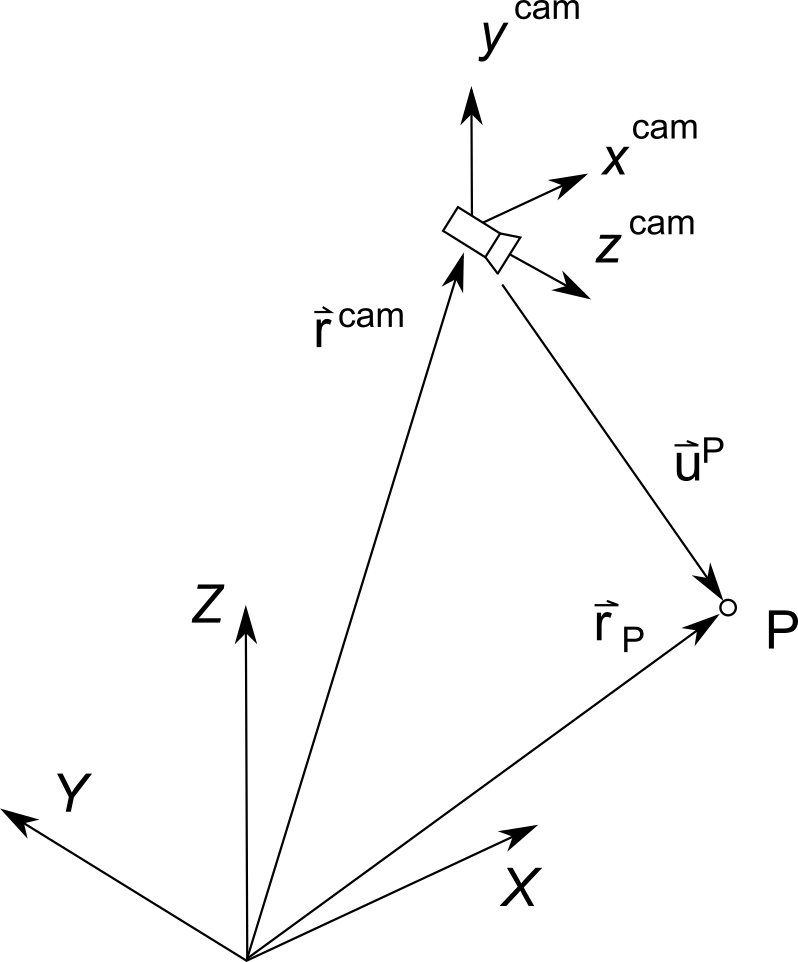}
	\caption{Camera position to world position}
	\label{fig:cinemWorldCam}
\end{figure}

Assume that the position and orientation of the camera with respect to a \textit{world frame} $\left\langle {X,\,\,Y,\,\,Z} \right\rangle $ are given by the position vector $\bf{r}^{cam}$ resolved in a world frame, and the transformation matrix $\bf{A}^{cam}$, respectively (see Fig. \ref{fig:cinemWorldCam}). The position vector of point $P$ in the world frame is given by: 

\begin{equation} \label{rP}
{{\bf{r}}^P} = {{\bf{r}}^{cam}} + {{\bf{A}}^{cam}}{{\bf{\bar u}}^P}
\end{equation}

This equation can be inverted to get the position of the point with respect to the camera frame ${{\bf{\bar u}}^P}$:

\begin{equation} \label{uP}
{{\bf{\bar u}}^P} = {\left( {{{\bf{A}}^{cam}}} \right)^T}\left( {{{\bf{r}}^P} - {{\bf{r}}^{cam}}} \right)
\end{equation}

As it is common in the kinematics used in robotics, this expression can be written in a more compact form using the $3\times4$ rigid body transformation matrix, as follows:

\begin{equation} \label{uP2}
{{\bf{\bar u}}^P} = \left[ {\begin{array}{*{20}{c}}
{{{\left( {{{\bf{A}}^{cam}}} \right)}^T}}&{ - {{{\bf{\bar r}}}^{cam}}}
\end{array}} \right]\left[ {\begin{array}{*{20}{c}}
{{{\bf{r}}^P}}\\
1
\end{array}} \right] = {{\bf{M}}^{ext}}\left[ {\begin{array}{*{20}{c}}
{{{\bf{r}}^P}}\\
1
\end{array}} \right]
\end{equation}

where ${{\bf{\bar r}}^{cam}} = {\left( {{{\bf{A}}^{cam}}} \right)^T}{{\bf{r}}^{cam}}$ contains the components of the position vector of the camera in the camera frame and ${{\bf{M}}^{ext}}$ is the $3 \times 4$ \textit{matrix of extrinsic parameters} of the camera, that is a function of its position and orientation with respect to the real world.

\section{Relation of the position of points in the image and the position in the world}  \label{sec:ImageToWorld}

Substituting Eq. \ref{uP2} into Eq. \ref{nPprima2} yields:

\begin{equation} \label{nPprima3}
{\grave{\bf{n}}^{P'}} = {{\bf{M}}^{int}}{{\bf{M}}^{ext}}\left[ {\begin{array}{*{20}{c}}
{{{\bf{r}}^P}}\\
1
\end{array}} \right]
\end{equation}

that relates the position of points in the image and the position in the world. As explained with Eq. \ref{nPprima2}, this equation can only be used in one direction: to find  once  is known. Equation \ref{nPprima3} can also be written in the alternative form:

\begin{equation} \label{nPprima4}
c\left[ {\begin{array}{*{20}{c}}
{{{\bf{n}}^{P'}}}\\
1
\end{array}} \right] = {{\bf{M}}^{int}}{{\bf{M}}^{ext}}\left[ {\begin{array}{*{20}{c}}
{{{\bf{r}}^P}}\\
1
\end{array}} \right]
\end{equation}

where $c$ is an unknown \textit{scale factor}. The matrix product on the right-hand side of this equation is called projection matrix ${\bf{P}} = {{\bf{M}}^{int}}{{\bf{M}}^{ext}}$.

\section{Motion tracking using computer vision}  \label{sec:motionTracking}

As explained above, motion tracking of a point $P$ cannot be done using a single video camera. One exception occurs when the point $P$ moves on a surface whose equation is known in the world frame. Assume that the equation of the surface takes the general intrinsic form $f(x, y, z) = 0$. Augmenting Eq. \ref{nPprima4} with this equation results in the following system of algebraic equations:

\begin{equation} \label{track_f}
\left\{ {\begin{array}{*{20}{c}}
{c\left[ {\begin{array}{*{20}{c}}
{{{\bf{n}}^{P'}}}\\
1
\end{array}} \right] - {{\bf{M}}^{int}}{{\bf{M}}^{ext}}\left[ {\begin{array}{*{20}{c}}
{{{\bf{r}}^P}}\\
1
\end{array}} \right] = {\bf{0}}}\\
{f\left( {r_x^P,r_y^P,r_z^P} \right) = 0}
\end{array}} \right.\,
\end{equation}

This is a system with 4 equations with 4 unknowns: the tree components of $\bf{r}^P$ and the scale factor $c$. A particular case of this situation happens when point $P$ moves on a plane. In this case Eq. \ref{track_f} yields: 

\begin{equation} \label{track_plane}
\left\{ {\begin{array}{*{20}{c}}
{c\left[ {\begin{array}{*{20}{c}}
{{{\bf{n}}^{P'}}}\\
1
\end{array}} \right] - {{\bf{M}}^{int}}{{\bf{M}}^{ext}}\left[ {\begin{array}{*{20}{c}}
{{{\bf{r}}^P}}\\
1
\end{array}} \right] = {\bf{0}}}\\
{Ar_x^P + Br_y^P + Cr_z^P + D = 0}
\end{array}} \right.\,
\end{equation}

where $A$, $B$, $C$, $D$ are the constants that define the plane. If the plane is, for example, the $\left\langle {X,\,\,Y} \right\rangle $ plane, Eq. \ref{track_plane} reduces to: 

\begin{equation} \label{track_planeXY}
c\left[ {\begin{array}{*{20}{c}}
{{{\bf{n}}^{P'}}}\\
1
\end{array}} \right] - {{\bf{M}}^{int}}{{\bf{M}}^{ext}}\left[ {\begin{array}{*{20}{c}}
{r_x^P}\\
{r_y^P}\\
0\\
1
\end{array}} \right] = {\bf{0}}
\end{equation}

that contains only 3 equations and 3 unknowns.

In the case that two cameras observe point $P$, say camera 1 and camera 2, motion tracking is always possible. Calculation of the position of $P$ in the world frame requires the solution of the system of equations:

\begin{equation} \label{track_2cam}
\left\{ {\begin{array}{*{20}{c}}
{{c_1}\left[ {\begin{array}{*{20}{c}}
{{\bf{n}}_1^{P'}}\\
1
\end{array}} \right] - {\bf{M}}_1^{int}{\bf{M}}_1^{ext}\left[ {\begin{array}{*{20}{c}}
{{{\bf{r}}^P}}\\
1
\end{array}} \right] = {\bf{0}}}\\
{{c_2}\left[ {\begin{array}{*{20}{c}}
{{\bf{n}}_2^{P'}}\\
1
\end{array}} \right] - {\bf{M}}_2^{int}{\bf{M}}_2^{ext}\left[ {\begin{array}{*{20}{c}}
{{{\bf{r}}^P}}\\
1
\end{array}} \right] = {\bf{0}}}
\end{array}} \right.\,
\end{equation}

where subscripts 1 and 2 are related to the parameters of the cameras or position in the images of each camera. Equation \ref{track_2cam} is an overdetermined (but compatible) linear system of 6 equations with 5 unknowns ($c_1$, $c_2$ and ${{\bf{r}}^{P}}$). These equations can be solved using a minimum-squared error procedure based on the pseudo-inverse of the coefficient matrix. 

Anyway, motion tracking can be performed once the cameras are calibrated. This is, once ${{\bf{M}}^{int}}$ and ${{\bf{M}}^{ext}}$ are found for each of the cameras. This procedure is called camera calibration and it is explained next.

\section{Camera calibration using Zhang method}  \label{sec:cameraCalibration}

The method presented by Zhang [2] calculates the value of the matrices ${{\bf{M}}^{int}}$ and ${{\bf{M}}^{ext}}$ for a given camera. In the version used here, a calibration pattern as the one shown in Fig. 4 is used. The pattern is a set of three chessboard prints forming a rectangular trihedral. The vertex of the trihedral is assumed to be the origin of the world frame $\left\langle {X,\,\,Y,\,\,Z} \right\rangle $ and the horizontal, back and lateral panels of the bookshelf, the $X-Y$, $Y-Z$ and $X-Z$ planes, respectively. 

\begin{figure}[htbp!]
	\centering
	\includegraphics[width=0.7\linewidth]{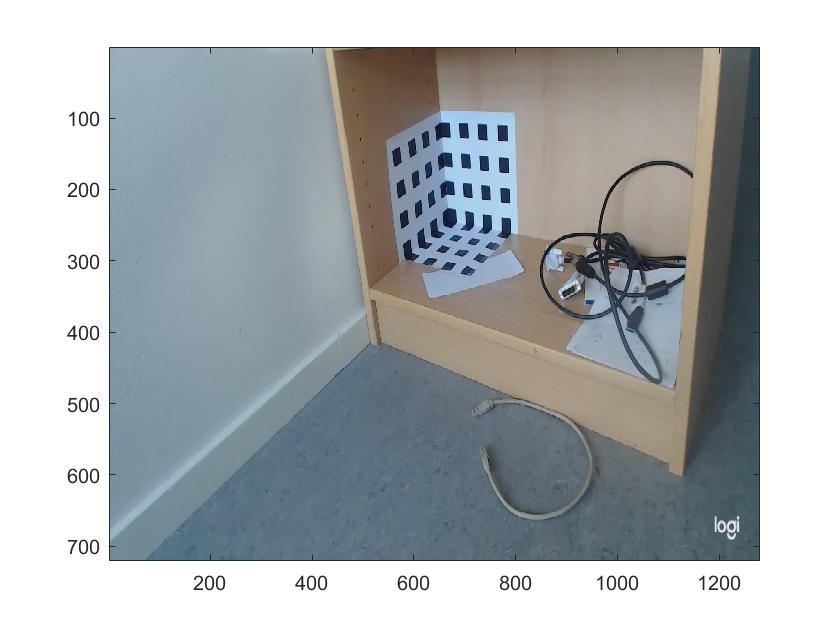}
	\caption{Calibration pattern used for camera calibration}
	\label{fig:calibPattern}
\end{figure}

Zhangs’ method requires as input data the position ${{\bf{r}}^P}$ of a set of points in the world frame and the corresponding position of them   ${{\bf{n}}^{P'}}$ in the image plane. In the example used in this document, 10 points are used in each of the $X-Y$, $Y-Z$ and $X-Z$ planes, resulting in a total of 30 calibration points. Figure \label{fig:calibPointsImage} shows the position of the points in the image plane while Fig. \label{fig:calibPointsWorld} shows them in the world frame. Table 1 shows the numerical values of the components of these vectors. Note that, as stated previously, in Fig. 4 the origin of $\left\langle {x^{im},\,\,y^{im}} \right\rangle $ is at the top-left corner and the positive direction of $y^{im}$ points downwards, while in Fig. 5-right the origin of $\left\langle {x^{im},\,\,y^{im}} \right\rangle $ is at the down-left corner and the positive direction of $y^{im}$ points upwards (as the axis are commonly represented).

\begin{figure}[htbp!]
	\centering
	\includegraphics[width=0.6\linewidth]{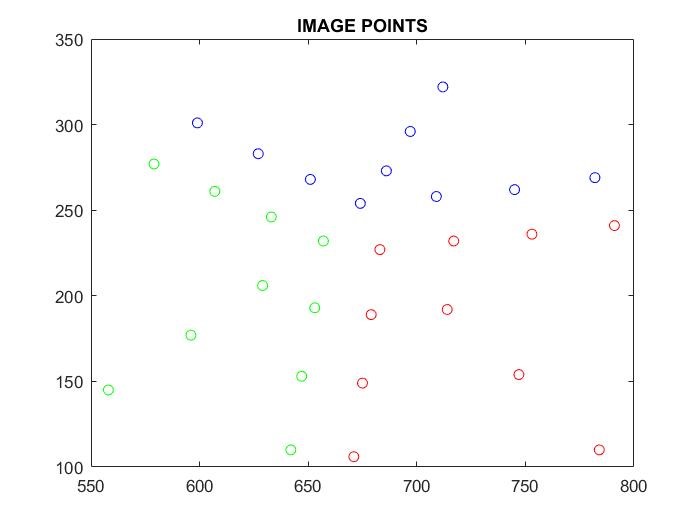}
	\caption{Calibration points in the image frame}
	\label{fig:calibPointsImage}
\end{figure}

\begin{figure}[htbp!]
	\centering
	\includegraphics[width=0.6\linewidth]{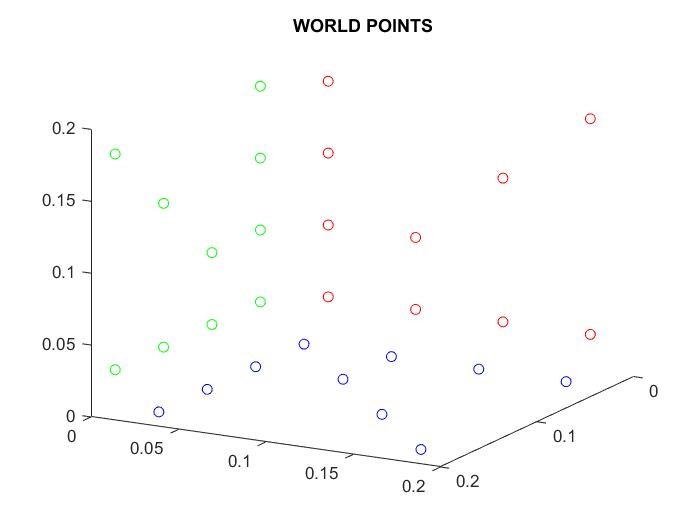}
	\caption{Calibration points in the world frame}
	\label{fig:calibPointsWorld}
\end{figure}

\begin{table}[h!]
\centering
\begin{tabular}{|c|c|c|c|c|c|c|c|c|c|c|c|c|c|c|}
	\hline
	\multicolumn{5}{|c|}{Plane $X-Y$} & \multicolumn{5}{|c|}{Plane $X-Z$} & \multicolumn{5}{|c|}{Plane $Y-Z$}  \\
	\hline
	\multicolumn{2}{|c|}{${{\bf{n}}^{P'}}$ (pix.)} & \multicolumn{3}{|c|}{${{\bf{r}}^P}$ (mm)}& \multicolumn{2}{|c|}{${{\bf{n}}^{P'}}$ (pix.)} & \multicolumn{3}{|c|}{${{\bf{r}}^P}$ (mm)} & \multicolumn{2}{|c|}{${{\bf{n}}^{P'}}$ (pix.)} & \multicolumn{3}{|c|}{${{\bf{r}}^P}$ (mm)}  \\
	\hline
	674 & 254 & 25 & 25 & 0 & 657 & 232 & 25 & 0 & 25 & 683 & 227 & 0 & 25 & 25 \\
	\hline
	686 & 273 & 75 & 75 & 0 & 629 & 206 & 75 & 0 & 75 & 714 & 192 & 0 & 75 & 75 \\
	\hline
	697 & 296 & 125 & 125 & 0 & 596 & 177 & 125 & 0 & 125 & 747 & 154 & 0 & 125 & 125 \\
	\hline
	712 & 322 & 175 & 175 & 0 & 558 & 145 & 175 & 0 & 175 & 784 & 110 & 0 & 175 & 175 \\
	\hline
	709 & 258 & 25 & 75 & 0 & 653 & 193 & 25 & 0 & 75 & 679 & 189 & 0 & 25 & 75 \\
	\hline
	745 & 262 & 25 & 125 & 0 & 647 & 153 & 25 & 0 & 125 & 675 & 149 & 0 & 25 & 125 \\
	\hline
	782 & 269 & 25 & 175 & 0 & 642 & 110 & 25 & 0 & 175 & 671 & 106 & 0 & 25 & 175 \\
	\hline
	651 & 268 & 75 & 25 & 0 & 633 & 246 & 75 & 0 & 25 & 717 & 232 & 0 & 75 & 25 \\
	\hline
	627 & 283 & 125 & 25 & 0 & 607 & 261 & 125 & 0 & 25 & 753 & 236 & 0 & 125 & 25 \\
	\hline
	599 & 301 & 175 & 25 & 0 & 579 & 277 & 175 & 0 & 25 & 791 & 241 & 0 & 175 & 25 \\
	\hline
\end{tabular}
\caption {Position vectors of calibration points in the image and world frames} \label{table:positionVectors}
\end{table}

Once the calibration points are selected and their coordinates are measured, Zhang’s calibration method follows 4 steps:

\begin{enumerate}
	\item Finding the homography matrices
	\item Finding the intrinsic parameters
	\item Finding the extrinsic parameters
	\item Optimizing intrinsic and extrinsic parameters
\end{enumerate}

These four steps are explained next.

\section{Finding the homography matrices}  \label{sec:Homography}

The application of Eq. \ref{nPprima4} to a set of points contained in a plane results in a plane-to-plane coordinate transformation, also called homography. Because the first 3 columns of $\bf{M}^\textit{ext}$ represents a rotation matrix, they can be interpreted as the components of the unitary vectors of the world frame in the camera frame, as follows:

\begin{equation} \label{Mext}
{{\bf{M}}^{ext}} = \left[ {\begin{array}{*{20}{c}}
{{{\left( {{{\bf{A}}^{cam}}} \right)}^T}}&{ - {{{\bf{\bar r}}}^{cam}}}
\end{array}} \right] = \left[ {\begin{array}{*{20}{c}}
{{{{\bf{\bar i}}}^{wd}}}&{{{{\bf{\bar j}}}^{wd}}}&{{{{\bf{\bar k}}}^{wd}}}&{ - {{{\bf{\bar r}}}^{cam}}}
\end{array}} \right]
\end{equation}

where superscript \textit{wd} stands for world frame. The application of Eq. \ref{nPprima4} to the $X-Y$ points yields:

\begin{equation} \label{pXY}
c\left[ {\begin{array}{*{20}{c}}
{{{\bf{n}}^{P'}}}\\
1
\end{array}} \right] = {{\bf{M}}^{int}}\left[ {\begin{array}{*{20}{c}}
{{{{\bf{\bar i}}}^{wd}}}&{{{{\bf{\bar j}}}^{wd}}}&{{{{\bf{\bar k}}}^{wd}}}&{ - {{{\bf{\bar r}}}^{cam}}}
\end{array}} \right]\left[ {\begin{array}{*{20}{c}}
{r_x^P}\\
{r_y^P}\\
0\\
1
\end{array}} \right] = {{\bf{M}}^{int}}\left[ {\begin{array}{*{20}{c}}
{{{{\bf{\bar i}}}^{wd}}}&{{{{\bf{\bar j}}}^{wd}}}&{ - {{{\bf{\bar r}}}^{cam}}}
\end{array}} \right]\left[ {\begin{array}{*{20}{c}}
{r_x^P}\\
{r_y^P}\\
1
\end{array}} \right]
\end{equation}

Calling homography matrix to the following $3 \times 3$ matrix: 

\begin{equation} \label{HXY}
{{\bf{H}}^{XY}} = {{\bf{M}}^{int}}\left[ {\begin{array}{*{20}{c}}
{{{{\bf{\bar i}}}^{wd}}}&{{{{\bf{\bar j}}}^{wd}}}&{ - {{{\bf{\bar r}}}^{cam}}}
\end{array}} \right]
\end{equation}

Equation \ref{pXY} can be written as:

\begin{equation} \label{eq24}
c\left[ {\begin{array}{*{20}{c}}
{n_x^{P'}}\\
{n_y^{P'}}\\
1
\end{array}} \right] = \left[ {\begin{array}{*{20}{c}}
{H_{11}^{XY}}&{H_{12}^{XY}}&{H_{13}^{XY}}\\
{H_{21}^{XY}}&{H_{22}^{XY}}&{H_{23}^{XY}}\\
{H_{31}^{XY}}&{H_{32}^{XY}}&{H_{33}^{XY}}
\end{array}} \right]\left[ {\begin{array}{*{20}{c}}
{r_x^P}\\
{r_y^P}\\
1
\end{array}} \right]
\end{equation}

Because there are 10 $P$-points in plane $X-Y$, 10 versions of Eq. \ref{eq24} can be written. For each point $P$, the scaling factor can be extracted from the 3rd equation, as follows: 

\begin{equation} \label{eq25}
c = H_{31}^{XY}r_x^P + H_{32}^{XY}r_y^P + H_{33}^{XY}
\end{equation}

Substituting this result in the $1^{st}$ and $2^{nd}$ equation of Eq. \ref{eq24} and rearranging yields:

\begin{equation} \label{eq26}
\begin{array}{l}
\left( {H_{31}^{XY}r_x^P + H_{32}^{XY}r_y^P + H_{33}^{XY}} \right)n_x^{P'} - \left( {H_{11}^{XY}r_x^P + H_{12}^{XY}r_y^P + H_{13}^{XY}} \right) = 0\\
\left( {H_{31}^{XY}r_x^P + H_{32}^{XY}r_y^P + H_{33}^{XY}} \right)n_y^{P'} - \left( {H_{21}^{XY}r_x^P + H_{22}^{XY}r_y^P + H_{23}^{XY}} \right) = 0
\end{array}
\end{equation}

Considering the elements of the homography matrix as the unknowns of a set of two linear homogeneous equations, Eq. \ref{eq26} can be written as: 

\begin{equation} \label{eq27}
\left[ {\begin{array}{*{20}{c}}
{ - r_x^P}&{ - r_y^P}&{ - 1}&0&0&0&{n_x^{P'}r_x^P}&{n_x^{P'}r_y^P}&{n_x^{P'}}\\
0&0&0&{ - r_x^P}&{ - r_y^P}&{ - 1}&{n_y^{P'}r_x^P}&{n_y^{P'}r_y^P}&{n_y^{P'}}
\end{array}} \right]\left[ {\begin{array}{*{20}{c}}
{H_{11}^{XY}}\\
{H_{12}^{XY}}\\
{H_{13}^{XY}}\\
{H_{21}^{XY}}\\
{H_{22}^{XY}}\\
{H_{23}^{XY}}\\
{H_{31}^{XY}}\\
{H_{32}^{XY}}\\
{H_{33}^{XY}}
\end{array}} \right] = {\bf{L}}_i^{XY}\left( {{{\bf{n}}^{P'}},{{\bf{r}}^P}} \right){{\bf{\hat H}}^{XY}} = {\bf{0}}
\end{equation}

where ${\bf{L}}_i^{XY}\left( {{{\bf{n}}^{P'}},{{\bf{r}}^P}} \right)$ is a $2 \times 9$ matrix that depends on the position of point $P$ in the image and the world frames and ${{\bf{\hat H}}^{XY}}$ is a $9 \times 1$ matrix that contains the elements of the homography matrix ${{\bf{H}}^{XY}}$ . Clearly, a set of two homogeneous equations can be obtained for each of the points P in the $X-Y$ plane. Gathering all the resulting equations yields:

\begin{equation} \label{eq28}
\left[ {\begin{array}{*{20}{c}}
{{\bf{L}}_1^{XY}}\\
{{\bf{L}}_2^{XY}}\\
 \vdots \\
{{\bf{L}}_{nXY}^{XY}}
\end{array}} \right]{{\bf{\hat H}}^{XY}} = {{\bf{L}}^{XY}}{{\bf{\hat H}}^{XY}} = {\bf{0}}
\end{equation}

where  is a $2nXY \times 9$ matrix and $nXY$ is the number of points used for calibration in the $X-Y$ plane ($nXY = 10$ in the example being used). Because the system of linear equations Eq. \ref{eq28} is homogeneous, non-trivial solutions for ${{\bf{\hat H}}^{XY}}$ can be found if the system is overdetermined (more equations than unknowns, this is $2nXY > 9\Rightarrow nXY > 5$). In that case, it can be demonstrated that the optimal solution to Eq. \ref{eq28} using minimum sum of squared errors as criterion is the eigenvector associated with the smallest eigenvalue of the matrix ${\left( {{{\bf{L}}^{XY}}} \right)^T}{{\bf{L}}^{XY}}$, this is:

\begin{equation} \label{eq29}
\begin{array}{l}
\left[ {{\bf{\lambda }},\bf{\phi }} \right] = {\rm{eig}}\left( {{{\left( {{{\bf{L}}^{XY}}} \right)}^T}{{\bf{L}}^{XY}}} \right);\\
\left[ {{\lambda _{\min }},{i_{\min }}} \right] = \min \left( {\bf{\lambda }} \right);\\
{{{\bf{\hat H}}}^{XY}} = \bf{\phi }\left( {:,{i_{\min }}} \right);
\end{array}
\end{equation}

where a Matlab-like nomenclature has been used to find the optimum value of ${{\bf{\hat H}}^{XY}}$. Because the system Eq. \ref{eq28} is homogeneous, with this procedure matrix ${{\bf{H}}^{XY}}$ is found up to a scale factor. This is, $e{{\bf{\hat H}}^{XY}}$ is also a solution of Eq. \ref{eq28}, being e any real number.

Clearly, the process Eq. \ref{pXY} – \ref{eq29} can be repeated for the points in planes $X-Z$ and $Y-Z$ to find the homography matrices ${{\bf{H}}^{XZ}}$ and ${{\bf{H}}^{YZ}}$.

\section{Finding the intrinsic parameters}  \label{sec:Intrinsics}

From Eq. \ref{HXY} it can be deduced that the unit vectors ${{\bf{\bar i}}^{wd}}$ and ${{\bf{\bar j}}^{wd}}$ are the first and second columns of the matrix ${\left( {{{\bf{M}}^{int}}} \right)^{ - 1}}{{\bf{H}}^{XY}}$, this is:

\begin{equation} \label{eq30}
\begin{array}{l}
{{{\bf{\bar i}}}^{wd}} = {\left( {{{\bf{M}}^{int}}} \right)^{ - 1}}{\bf{h}}_1^{XY},\\
{{{\bf{\bar j}}}^{wd}} = {\left( {{{\bf{M}}^{int}}} \right)^{ - 1}}{\bf{h}}_2^{XY},
\end{array}
\end{equation}

where ${\bf{h}}_i^{XY}$ is the ith column of ${{\bf{H}}^{XY}}$. Because ${{\bf{\bar i}}^{wd}}$ and ${{\bf{\bar j}}^{wd}}$ are unit and orthogonal vectors, the following equations can be deduced: 

\begin{equation} \label{eq31}
\left\{ {\begin{array}{*{20}{c}}
{{{\left( {{\bf{h}}_1^{XY}} \right)}^T}{{\left( {{{\bf{M}}^{int}}} \right)}^{ - T}}{{\left( {{{\bf{M}}^{int}}} \right)}^{ - 1}}{\bf{h}}_2^{XY} = 0}\\
{{{\left( {{\bf{h}}_1^{XY}} \right)}^T}{{\left( {{{\bf{M}}^{int}}} \right)}^{ - T}}{{\left( {{{\bf{M}}^{int}}} \right)}^{ - 1}}{\bf{h}}_1^{XY} - {{\left( {{\bf{h}}_2^{XY}} \right)}^T}{{\left( {{{\bf{M}}^{int}}} \right)}^{ - T}}{{\left( {{{\bf{M}}^{int}}} \right)}^{ - 1}}{\bf{h}}_2^{XY} = 0}
\end{array}} \right.
\end{equation}

where ${\left( {{{\bf{M}}^{int}}} \right)^{ - T}} = {\left( {{{\left( {{{\bf{M}}^{int}}} \right)}^{ - 1}}} \right)^T}$. Matrix  ${\bf{B}}$ is defined as follows:

\begin{equation} \label{eq32}
{\bf{B}} = {\left( {{{\bf{M}}^{int}}} \right)^{ - T}}{\left( {{{\bf{M}}^{int}}} \right)^{ - 1}} = \left[ {\begin{array}{*{20}{c}}
{\frac{1}{{{\alpha ^2}}}}&{ - \frac{\gamma }{{{\alpha ^2}\beta }}}&{\frac{{n_y^O\gamma  - n_x^O\beta }}{{{\alpha ^2}\beta }}}\\
{}&{\frac{\gamma }{{{\alpha ^2}{\beta ^2}}} + \frac{1}{{{\beta ^2}}}}&{ - \frac{{\gamma \left( {n_y^O\gamma  - n_x^O\beta } \right)}}{{{\alpha ^2}{\beta ^2}}} - \frac{{n_y^O}}{{{\beta ^2}}}}\\
{symmetric}&{}&{\frac{{{{\left( {n_y^O\gamma  - n_x^O\beta } \right)}^2}}}{{{\alpha ^2}{\beta ^2}}} + \frac{{{{\left( {n_y^O} \right)}^2}}}{{{\beta ^2}}} + 1}
\end{array}} \right]
\end{equation}

This matrix has been explicitly calculated in terms of the elements of given in Eq. \ref{Mint2}.${\bf{B}}$ is a symmetric matrix. Equation \ref{eq31} includes three matrix products like:

\begin{equation} \label{eq33}
\begin{array}{l}
{\left( {{\bf{h}}_i^{XY}} \right)^T}{\bf{Bh}}_i^{XY} = \\
\left[ {\begin{array}{*{20}{c}}
{h_{i1}^{XY}h_{j1}^{XY}}&{h_{i1}^{XY}h_{j2}^{XY} + h_{i2}^{XY}h_{j1}^{XY}}&{h_{i2}^{XY}h_{j2}^{XY}}&{h_{i3}^{XY}h_{j1}^{XY} + h_{i1}^{XY}h_{j3}^{XY}}&{h_{i3}^{XY}h_{j2}^{XY} + h_{i2}^{XY}h_{j3}^{XY}}&{h_{i3}^{XY}h_{j3}^{XY}}
\end{array}} \right]\left[ {\begin{array}{*{20}{c}}
{{B_{11}}}\\
{{B_{12}}}\\
{{B_{22}}}\\
{{B_{13}}}\\
{{B_{23}}}\\
{{B_{33}}}
\end{array}} \right] = \\
 = {\bf{v}}_{ij}^{XY}{\bf{\hat B}}
\end{array}
\end{equation}

Where ${\bf{v}}_{ij}^{XY}$ is a $1 \times 6$ matrix that can be built using the elements of ${{\bf{H}}^{XY}}$  and ${\bf{\hat B}}$ is a $6 \times 1$ matrix that contains the elements of ${\bf{B}}$. Using this nomenclature, the two equations in Eq. \ref{eq31} can be written as: 

\begin{equation} \label{eq34}
\left[ {\begin{array}{*{20}{c}}
{{\bf{v}}_{12}^{XY}}\\
{{\bf{v}}_{11}^{XY} - {\bf{v}}_{22}^{XY}}
\end{array}} \right]{\bf{\hat B}} = {\bf{0}}
\end{equation}

From the homography matrices ${{\bf{H}}^{XZ}}$ and ${{\bf{H}}^{YZ}}$ , 4 more equations like those given in Eq. \ref{eq31} can be written. Adding these equations to the set given in Eq. \ref{eq34} yields:

\begin{equation} \label{eq35}
\left[ {\begin{array}{*{20}{c}}
{{\bf{v}}_{12}^{XY}}\\
{{\bf{v}}_{11}^{XY} - {\bf{v}}_{22}^{XY}}\\
{{\bf{v}}_{12}^{XZ}}\\
{{\bf{v}}_{11}^{XZ} - {\bf{v}}_{22}^{XZ}}\\
{{\bf{v}}_{12}^{YZ}}\\
{{\bf{v}}_{11}^{YZ} - {\bf{v}}_{22}^{YZ}}
\end{array}} \right]{\bf{\hat B}} = {\bf{V\hat B}} = {\bf{0}}
\end{equation}

where  ${\bf{V}}$ is a $6 \times 6$ matrix that can be built as a function of the three homography matrices.  Equation \ref{eq35} is again a system of linear-homogeneous equation that can be solved using the same procedure as before to find ${{\bf{\hat H}}^{XY}}$. In this case the unknown ${\bf{\hat B}}$ is obtained as:

\begin{equation} \label{eq36}
\begin{array}{l}
\left[ {{\bf{\lambda }},{\phi }} \right] = {\rm{eig}}\left( {{{\bf{V}}^T}{\bf{V}}} \right);\\
\left[ {{\lambda _{\min }},{i_{\min }}} \right] = \min \left( {\bf{\lambda }} \right);\\
{\bf{\hat B}} = {\phi }\left( {:,{i_{\min }}} \right);
\end{array}
\end{equation}

The last step to find ${{{\bf{M}}^{int}}}$ is to get its components out of $\bf{B}$. This task can be done using the following explicit formulas:

\begin{equation} \label{eq37}
\begin{array}{l}
n_y^O = {{\left( {{B_{12}}{B_{13}} - {B_{11}}{B_{23}}} \right)} \mathord{\left/
 {\vphantom {{\left( {{B_{12}}{B_{13}} - {B_{11}}{B_{23}}} \right)} {\left( {{B_{11}}{B_{22}} - {B_{12}}^2} \right)}}} \right.
 \kern-\nulldelimiterspace} {\left( {{B_{11}}{B_{22}} - {B_{12}}^2} \right)}}\\
\lambda  = {B_{33}} - {{\left( {{B_{13}}^2 + n_y^O\left( {{B_{12}}{B_{13}} - {B_{11}}{B_{23}}} \right)} \right)} \mathord{\left/
 {\vphantom {{\left( {{B_{13}}^2 + n_y^O\left( {{B_{12}}{B_{13}} - {B_{11}}{B_{23}}} \right)} \right)} {{B_{11}}}}} \right.
 \kern-\nulldelimiterspace} {{B_{11}}}}\\
\alpha  =  - \sqrt {{\lambda  \mathord{\left/
 {\vphantom {\lambda  {{B_{11}}}}} \right.
 \kern-\nulldelimiterspace} {{B_{11}}}}} \\
\beta  =  - \sqrt {{{\lambda {B_{11}}} \mathord{\left/
 {\vphantom {{\lambda {B_{11}}} {\left( {{B_{11}}{B_{22}} - {B_{12}}^2} \right)}}} \right.
 \kern-\nulldelimiterspace} {\left( {{B_{11}}{B_{22}} - {B_{12}}^2} \right)}}} \\
\gamma  =  - {{{B_{12}}^2{\alpha ^2}\beta } \mathord{\left/
 {\vphantom {{{B_{12}}^2{\alpha ^2}\beta } \lambda }} \right.
 \kern-\nulldelimiterspace} \lambda }\\
n_x^O = {{\gamma n_y^O} \mathord{\left/
 {\vphantom {{\gamma n_y^O} \beta }} \right.
 \kern-\nulldelimiterspace} \beta } - {B_{13}}{{{\alpha ^2}} \mathord{\left/
 {\vphantom {{{\alpha ^2}} \lambda }} \right.
 \kern-\nulldelimiterspace} \lambda }
\end{array}
\end{equation}

\section{Finding the extrinsic parameters}  \label{sec:Extrinsics}

From Eq. \ref{HXY}, the unit vectors can be obtained as:

\begin{equation} \label{eq38}
\begin{array}{l}
{\bf{\bar i}}_{XY}^{wd} = e{\left( {{{\bf{M}}^{int}}} \right)^{ - 1}}{\bf{h}}_1^{XY}\\
{\bf{\bar j}}_{XY}^{wd} = e{\left( {{{\bf{M}}^{int}}} \right)^{ - 1}}{\bf{h}}_2^{XY}
\end{array}
\end{equation}

where subscripts $XY$ means that these are the unit vectors computed form the $XY$ homography and $e$ is the scale factor that remains unknowns for the homography matrix as stated in Section \ref{sec:Homography}. Because these are unit vectors, the following values for $e$ can be obtained:

\begin{equation} \label{eq39}
\begin{array}{l}
{e_1} = {1 \mathord{\left/
 {\vphantom {1 {\left\| {{{\left( {{{\bf{M}}^{int}}} \right)}^{ - 1}}{\bf{h}}_1^{XY}} \right\|}}} \right.
 \kern-\nulldelimiterspace} {\left\| {{{\left( {{{\bf{M}}^{int}}} \right)}^{ - 1}}{\bf{h}}_1^{XY}} \right\|}}\\
{e_2} = {1 \mathord{\left/
 {\vphantom {1 {\left\| {{{\left( {{{\bf{M}}^{int}}} \right)}^{ - 1}}{\bf{h}}_2^{XY}} \right\|}}} \right.
 \kern-\nulldelimiterspace} {\left\| {{{\left( {{{\bf{M}}^{int}}} \right)}^{ - 1}}{\bf{h}}_2^{XY}} \right\|}}
\end{array}
\end{equation}

Ideally $e = {e_1} = {e_2}$. However, because ${{\bf{H}}^{XY}}$ is the result of an approximation (minimum sum of squared errors) in practice ${e_1} \ne {e_2}$. The unit vectors can be computed using the following formulas:

\begin{equation} \label{eq40}
\begin{array}{l}
{\bf{\bar i}}_{XY}^{wd} = {{{{\left( {{{\bf{M}}^{int}}} \right)}^{ - 1}}{\bf{h}}_1^{XY}} \mathord{\left/
 {\vphantom {{{{\left( {{{\bf{M}}^{int}}} \right)}^{ - 1}}{\bf{h}}_1^{XY}} {\left\| {{{\left( {{{\bf{M}}^{int}}} \right)}^{ - 1}}{\bf{h}}_1^{XY}} \right\|}}} \right.
 \kern-\nulldelimiterspace} {\left\| {{{\left( {{{\bf{M}}^{int}}} \right)}^{ - 1}}{\bf{h}}_1^{XY}} \right\|}}\\
{\bf{\bar j}}_{XY}^{wd} = {{{{\left( {{{\bf{M}}^{int}}} \right)}^{ - 1}}{\bf{h}}_2^{XY}} \mathord{\left/
 {\vphantom {{{{\left( {{{\bf{M}}^{int}}} \right)}^{ - 1}}{\bf{h}}_2^{XY}} {\left\| {{{\left( {{{\bf{M}}^{int}}} \right)}^{ - 1}}{\bf{h}}_2^{XY}} \right\|}}} \right.
 \kern-\nulldelimiterspace} {\left\| {{{\left( {{{\bf{M}}^{int}}} \right)}^{ - 1}}{\bf{h}}_2^{XY}} \right\|}}\\
{\bf{\bar k}}_{XY}^{wd} = {\bf{\bar i}}_{XY}^{wd} \times {\bf{\bar j}}_{XY}^{wd}
\end{array}
\end{equation}

and the position vector of the origin of the camera frame is found as:

\begin{equation} \label{eq41}
{\bf{\bar r}}_{XY}^{cam} =  - e{\left( {{{\bf{M}}^{int}}} \right)^{ - 1}}{\bf{h}}_3^{XY}\, =  - \frac{{{e_1} + {e_2}}}{2}{\left( {{{\bf{M}}^{int}}} \right)^{ - 1}}{\bf{h}}_3^{XY}
\end{equation}

where $e$ is approximated as the average of $e_1$ and $e_2$ previously defined. Using these results, the matrix of extrinsic parameters can be obtained from the $XY$ homography as follows:

\begin{equation} \label{eq42}
{\bf{M}}_{XY}^{ext} = \left[ {\begin{array}{*{20}{c}}
{{\bf{\bar i}}_{XY}^{wd}}&{{\bf{\bar j}}_{XY}^{wd}}&{{\bf{\bar k}}_{XY}^{wd}}&{ - {\bf{\bar r}}_{XY}^{cam}}
\end{array}} \right] = \left[ {\begin{array}{*{20}{c}}
{{{\left( {{\bf{A}}_{XY}^{cam}} \right)}^T}}&{ - {\bf{\bar r}}_{XY}^{cam}}
\end{array}} \right]
\end{equation}

Because the homography ${{\bf{H}}^{XY}}$ is the result of an approximation, the resulting rotation matrix ${\bf{A}}_{XY}^{cam}$ is not exact, therefore, it does not fulfill the orthogonality property: ${\left( {{\bf{A}}_{XY}^{cam}} \right)^{ - 1}} = {\left( {{\bf{A}}_{XY}^{cam}} \right)^T}$ . There is a well-known procedure to find the best “true” rotation matrix to approximate the obtained ${\bf{A}}_{XY}^{cam}$. The procedure starts by computing the singular value decomposition of  . Then, the corrected is obtained as the product of the left-singular-vectors matrix times the right-singular-vectors matrix, as follows:

\begin{equation} \label{eq43}
\begin{array}{l}
\left[ {{\bf{U}},{\bf{S}},{\bf{V}}} \right] = {\rm{svd}}\left( {{\bf{A}}_{XY}^{cam}} \right);\\
{\left( {{\bf{A}}_{XY}^{cam}} \right)_{corrected}} = {\bf{UV}};
\end{array}
\end{equation}

Using similar reasoning, other two matrices of extrinsic parameters can be found using the homographies ${{\bf{H}}^{XZ}}$ and ${{\bf{H}}^{YZ}}$, as follows:

\begin{equation} \label{eq44}
\begin{array}{l}
{\bf{\bar i}}_{XZ}^{wd} = {{{{\left( {{{\bf{M}}^{int}}} \right)}^{ - 1}}{\bf{h}}_1^{XZ}} \mathord{\left/
 {\vphantom {{{{\left( {{{\bf{M}}^{int}}} \right)}^{ - 1}}{\bf{h}}_1^{XZ}} {\left\| {{{\left( {{{\bf{M}}^{int}}} \right)}^{ - 1}}{\bf{h}}_1^{XZ}} \right\|}}} \right.
 \kern-\nulldelimiterspace} {\left\| {{{\left( {{{\bf{M}}^{int}}} \right)}^{ - 1}}{\bf{h}}_1^{XZ}} \right\|}}\\
{\bf{\bar k}}_{XZ}^{wd} = {{{{\left( {{{\bf{M}}^{int}}} \right)}^{ - 1}}{\bf{h}}_2^{XZ}} \mathord{\left/
 {\vphantom {{{{\left( {{{\bf{M}}^{int}}} \right)}^{ - 1}}{\bf{h}}_2^{XZ}} {\left\| {{{\left( {{{\bf{M}}^{int}}} \right)}^{ - 1}}{\bf{h}}_2^{XZ}} \right\|}}} \right.
 \kern-\nulldelimiterspace} {\left\| {{{\left( {{{\bf{M}}^{int}}} \right)}^{ - 1}}{\bf{h}}_2^{XZ}} \right\|}}\\
{\bf{\bar j}}_{XZ}^{wd} = {\bf{\bar k}}_{XZ}^{wd} \times {\bf{\bar i}}_{XZ}^{wd}\\
{\bf{\bar r}}_{XZ}^{cam} =  - \frac{{{e_1} + {e_2}}}{2}{\left( {{{\bf{M}}^{int}}} \right)^{ - 1}}{\bf{h}}_3^{XZ}\\
\,{\bf{M}}_{XZ}^{ext} = \left[ {\begin{array}{*{20}{c}}
{{\bf{\bar i}}_{XZ}^{wd}}&{{\bf{\bar j}}_{XZ}^{wd}}&{{\bf{\bar k}}_{XZ}^{wd}}&{ - {\bf{\bar r}}_{XZ}^{cam}}
\end{array}} \right] = \left[ {\begin{array}{*{20}{c}}
{\left( {{\bf{A}}_{XZ}^{cam}} \right)_{corrected}^T}&{ - {\bf{\bar r}}_{XZ}^{cam}}
\end{array}} \right]
\end{array}
\end{equation}

\begin{equation} \label{eq45}
\begin{array}{l}
{\bf{\bar j}}_{YZ}^{wd} = {{{{\left( {{{\bf{M}}^{int}}} \right)}^{ - 1}}{\bf{h}}_1^{YZ}} \mathord{\left/
 {\vphantom {{{{\left( {{{\bf{M}}^{int}}} \right)}^{ - 1}}{\bf{h}}_1^{YZ}} {\left\| {{{\left( {{{\bf{M}}^{int}}} \right)}^{ - 1}}{\bf{h}}_1^{YZ}} \right\|}}} \right.
 \kern-\nulldelimiterspace} {\left\| {{{\left( {{{\bf{M}}^{int}}} \right)}^{ - 1}}{\bf{h}}_1^{YZ}} \right\|}}\\
{\bf{\bar k}}_{YZ}^{wd} = {{{{\left( {{{\bf{M}}^{int}}} \right)}^{ - 1}}{\bf{h}}_2^{YZ}} \mathord{\left/
 {\vphantom {{{{\left( {{{\bf{M}}^{int}}} \right)}^{ - 1}}{\bf{h}}_2^{YZ}} {\left\| {{{\left( {{{\bf{M}}^{int}}} \right)}^{ - 1}}{\bf{h}}_2^{YZ}} \right\|}}} \right.
 \kern-\nulldelimiterspace} {\left\| {{{\left( {{{\bf{M}}^{int}}} \right)}^{ - 1}}{\bf{h}}_2^{YZ}} \right\|}}\\
{\bf{\bar i}}_{YZ}^{wd} = {\bf{\bar j}}_{YZ}^{wd} \times {\bf{\bar k}}_{YZ}^{wd}\\
{\bf{\bar r}}_{YZ}^{cam} =  - \frac{{{e_1} + {e_2}}}{2}{\left( {{{\bf{M}}^{int}}} \right)^{ - 1}}{\bf{h}}_3^{YZ}\\
\,{\bf{M}}_{YZ}^{ext} = \left[ {\begin{array}{*{20}{c}}
{{\bf{\bar i}}_{YZ}^{wd}}&{{\bf{\bar j}}_{YZ}^{wd}}&{{\bf{\bar k}}_{YZ}^{wd}}&{ - {\bf{\bar r}}_{YZ}^{cam}}
\end{array}} \right] = \left[ {\begin{array}{*{20}{c}}
{\left( {{\bf{A}}_{YZ}^{cam}} \right)_{corrected}^T}&{ - {\bf{\bar r}}_{YZ}^{cam}}
\end{array}} \right]
\end{array}
\end{equation}

However, the matrix of extrinsic parameters ${{\bf{M}}^{ext}}$ is unique. Therefore, one must find an unified version out of the three results ${\bf{M}}_{XY}^{ext}$, ${\bf{M}}_{XZ}^{ext}$ and ${\bf{M}}_{YZ}^{ext}$. This calculation is performed in the last step of the calibration process.

\section{Unified value of the extrinsic parameters}  \label{sec:unifiedExt}

Parametrization of a rotation matrix can be done using different rotation coordinates as Euler angles or unit quaternions (also called Euler parameters). However, in this work, it is convenient to use the so-called \textit{Rodriguez parameters}. Rodriguez parameters are defined using the rotation angle $\alpha$ and the unit vector $\bf{e}$ (components in world frame) that are defined in \textit{Euler’s theorem of finite rotations} as the angle that the global frame (world frame) has to be rotated and the direction of this rotation to make it parallel to the moving frame (camera frame). Rodriguez parameters are defined as: 

\begin{equation} \label{eq46}
{{\bf{g}}^{cam}} = \tan \left( {\frac{\alpha }{2}} \right){\bf{e}}
\end{equation}

The rotation matrix can be obtained as a function of the Rodriguez parameters using the following formula: 

\begin{equation} \label{eq47}
{{\bf{A}}^{cam}} = {{\bf{1}}_{3 \times 3}} + \frac{2}{{1 + {{\left( {{{\bf{g}}^{cam}}} \right)}^T}{{\bf{g}}^{cam}}}}\left( {{{{\bf{\tilde g}}}^{cam}} + {{{\bf{\tilde g}}}^{cam}}{{{\bf{\tilde g}}}^{cam}}} \right)
\end{equation}

where ${{\bf{\tilde g}}^{cam}}$ is the skew-symmetric matrix associated with ${{\bf{g}}^{cam}}$. Equation \ref{eq47} can be inverted to find the Rodriguez parameters from the rotation matrix, as follows: 

\begin{equation} \label{eq48}
{{\bf{g}}^{cam}} = \frac{1}{{{\rm{trace}}\left( {{{\bf{A}}^{cam}}} \right)}}\left[ {\begin{array}{*{20}{c}}
{A_{23}^{cam} - A_{32}^{cam}}\\
{A_{31}^{cam} - A_{13}^{cam}}\\
{A_{12}^{cam} - A_{21}^{cam}}
\end{array}} \right]
\end{equation}

This formula can be used to find ${\bf{g}}_{XY}^{cam},\,\,{\bf{g}}_{XZ}^{cam},\,\,{\bf{g}}_{YZ}^{cam}$ using the matrices ${\bf{A}}_{XY}^{cam},\,\,{\bf{A}}_{XZ}^{cam},\,\,{\bf{A}}_{XZ}^{cam}$, respectively. 
Using Rodriguez parameters, the 12 elements of the matrix ${{\bf{M}}^{ext}}$ can be obtained as a function of 6 independent parameters: the 3 components of ${{\bf{\bar r}}^{cam}}$ and the 3 components of ${{\bf{g}}^{cam}}$. After the calculations described in the previous section, there are 3 possible values of ${{\bf{\bar r}}^{cam}}$ and ${{\bf{g}}^{cam}}$ resulting from the three homografies. An estimate of the “true” values can be obtained as an average, as follows:

\begin{equation} \label{eq49}
\begin{array}{l}
{\bf{g}}_0^{cam} = \frac{1}{3}\left( {{\bf{g}}_{XY}^{cam} + {\bf{g}}_{XZ}^{cam} + {\bf{g}}_{YZ}^{cam}} \right),\\
{\bf{\bar r}}_0^{cam} = \frac{1}{3}\left( {{\bf{\bar r}}_{XY}^{cam} + {\bf{\bar r}}_{XZ}^{cam} + {\bf{\bar r}}_{YZ}^{cam}} \right),
\end{array}
\end{equation}

where subscript ‘0’ means that these values are estimates of the optimized values. 

\section{Optimization of projection matrix }  \label{sec:Optimization}

The projection matrix ${\bf{P}} = {{\bf{M}}^{int}}{{\bf{M}}^{ext}}$ can be optimized for a given camera. As a result of the calculations presented in the previous sections, Eqs. \ref{nPprima3}-\ref{nPprima4} is satisfied only approximately for the set of points P used in the calibration process. Using Eq. \ref{nPprima3} as a reference, one can write:

\begin{equation} \label{eq50}
{\grave{\bf{n}}^{P'}} \simeq {{\bf{M}}^{int}}{{\bf{M}}^{ext}}\left[ {\begin{array}{*{20}{c}}
{{{\bf{r}}^P}}\\
1
\end{array}} \right] = {\grave{\hat{\bf{n}}}^{P'}}
\end{equation}

This is, the projected homogeneous vectors ${\grave{\hat{\bf{n}}}^{P'}}$ are not exactly equal to the image vectors homogeneous  ${\grave{\bf{n}}^{P'}}$. The error in the projection matrix associated with point P can be measured using the following norm: 

\begin{equation} \label{eq51}
{\varepsilon _P} = \left\| {{{\hom }^{ - 1}}\left( {{{\grave{\bf{n}}}^{P'}}} \right) - {{\hom }^{ - 1}}\left( {\grave{\hat{\bf{n}}}^{P'}} \right)} \right\|
\end{equation}

The projection matrix can be optimized to minimize the sum of the squared errors of all calibration points $P_i$. The sum of the squared errors is a function of the independent intrinsic and extrinsic parameters, as follows:

\begin{equation} \label{eq52}
E({{\bf{p}}^{cam}}) = \sum\limits_{i = 1}^{ncal} {{{\left( {{\varepsilon _{{P_i}}}} \right)}^2}} 
\end{equation}

where $ncal$ is the total number of calibration points and

\begin{equation} \label{eq53}
{{\bf{p}}^{cam}} = {\left[ {\begin{array}{*{20}{c}}
\alpha &\beta &\gamma &{n_x^O}&{n_y^O}&{{{\left( {{{\bf{g}}^{cam}}} \right)}^T}}&{{{\left( {{{{\bf{\bar r}}}^{cam}}} \right)}^T}}
\end{array}} \right]^T}
\end{equation}

contains the 11 independent parameters needed to find ${{\bf{M}}^{int}}$ and ${{\bf{M}}^{ext}}$. The optimum value of  ${{\bf{p}}^{cam}}$ is obtained with a nonlinear optimization process, as follows:

\begin{equation} \label{eq54}
{\bf{p}}_{opt}^{cam} = \min E({{\bf{p}}^{cam}})
\end{equation}

As initial guess of the optimization, the values of the intrinsic parameters given in Eq. \ref{eq37} and the values of the extrinsic parameters given in Eq. \ref{eq49} can be used.

\iffalse

\bibliographystyle{unsrt}  
%\bibliography{references}  %%% Remove comment to use the external .bib file (using bibtex).
%%% and comment out the ``thebibliography'' section.

\fi

%%% Comment out this section when you \bibliography{references} is enabled.

\end{document}